\documentclass{article}

\PassOptionsToPackage{numbers,sort&compress}{natbib}
\usepackage[preprint]{tackling_climate_workshop_style}
\makeatletter\renewcommand{\@noticestring}{Preprint.}\makeatother

\usepackage[utf8]{inputenc}
\usepackage[T1]{fontenc}

\usepackage{hyperref}
\hypersetup{hidelinks, pdftitle={Watts If? A Benchmark for Counterfactual Reasoning About the AI Data Center Buildout}}
\usepackage{orcidlink}
\usepackage{url}
\usepackage{booktabs}
\usepackage{amsmath}
\usepackage{amsfonts}
\usepackage{amssymb}
\usepackage{microtype}
\usepackage{graphicx}
\usepackage{tabularx}
\usepackage{tikz}
\usetikzlibrary{arrows.meta,positioning,fit,calc}

\graphicspath{{figures/}}

\newcommand{\code}[1]{\texttt{\small #1}}
\newcommand{\figmono}{\fontsize{7.2}{8.3}\selectfont\ttfamily}
\newcommand{\figtxt}{\fontsize{7.2}{8.3}\selectfont}
\newcommand{\fignum}{\fontsize{7}{8}\selectfont\bfseries}

\newcommand{\nGenQ}{200}
\newcommand{\nMcqQ}{240}

\newcommand{\nOpenDeploy}{11}
\newcommand{\nOpenWeights}{7}

\newcommand{\nMcqOpenRun}{9}
\newcommand{\GenOpenMin}{12.5}
\newcommand{\GenOpenMax}{71.5}
\newcommand{\McqOpenMin}{58.3}
\newcommand{\McqOpenMax}{92.5}
\newcommand{\GenHalfMin}{4.6}
\newcommand{\GenHalfMax}{6.9}
\newcommand{\McqHalfMin}{3.4}
\newcommand{\McqHalfMax}{6.2}
\newcommand{\GenTiedN}{six}
\newcommand{\McqTiedN}{five}

\newcommand{\GenTiedMin}{65.5}
\newcommand{\GenTiedMax}{71.5}
\newcommand{\McqTiedMin}{90.4}
\newcommand{\McqTiedMax}{92.5}
\newcommand{\ProvGapMin}{0.5}
\newcommand{\ProvGapMax}{3.0}
\newcommand{\ProvGapPMin}{0.36}
\newcommand{\QwenGap}{10.0}
\newcommand{\QwenGapP}{0.010}

\newcommand{\WeakGenMin}{12.5}
\newcommand{\WeakGenMax}{18.0}
\newcommand{\WeakMcqMin}{58.3}
\newcommand{\WeakMcqMax}{80.0}
\newcommand{\AppStrictMax}{4.0}
\newcommand{\AppZeroOpen}{10}
\newcommand{\AppNoopShare}{66}
\newcommand{\AppGoldMean}{20.5}
\newcommand{\AppPrecMin}{0.84}
\newcommand{\AppPrecMax}{0.94}

\newcommand{\AppRecChgMin}{72}
\newcommand{\AppRecChgMax}{85}
\newcommand{\AppRecNoopMin}{15}
\newcommand{\AppRecNoopMax}{31}
\newcommand{\AppOmitNoopMin}{85}
\newcommand{\AppOmitNoopMax}{91}
\newcommand{\AppExclMax}{24.0}
\newcommand{\AppJeightyMax}{12.0}

\newcommand{\AppJaccardMax}{0.47}
\newcommand{\AppFoneMax}{0.61}
\newcommand{\ProbeAppF}{0.71}
\newcommand{\ProbeAppJ}{0.56}

\newcommand{\ProbeAppJfifty}{80}
\newcommand{\ProbeGenMax}{12.0}
\newcommand{\McqLongest}{42.5}
\newcommand{\McqHardN}{138}
\newcommand{\McqHardStrongMin}{89.1}
\newcommand{\McqHardStrongMax}{91.3}
\newcommand{\McqHardDropMin}{2.5}
\newcommand{\McqHardDropMax}{6.7}

\newcommand{\ToolCompoundGen}{128}
\newcommand{\ToolCompoundMcq}{82}

\newcommand{\ToolCompoundMiniGen}{1}
\newcommand{\ToolCompoundMiniMcq}{3}

\newcommand{\CostOpenRouter}{2.00}
\newcommand{\AppMissChgMin}{15}
\newcommand{\AppMissChgMax}{28}
\newcommand{\McqAppMin}{75.9}
\newcommand{\McqAppMax}{96.6}
\newcommand{\AppStrictStrongMax}{0.0}
\newcommand{\WorldStates}{24{,}659}
\newcommand{\FmtStrongMax}{5}
\newcommand{\FmtWeakMax}{48}

\title{Watts If? A Benchmark for Counterfactual Reasoning\\
About the AI Data Center Buildout}

\author{%
  Pankaj Chouhan\,\thanks{Equal contribution.} \\
  \textbf{\small AlphaInfra} \\
  \texttt{\small pankaj.c@alphainfra.ai}
  \And
  Rajarshi Das\footnotemark[1] \\
  \textbf{\small AlphaInfra} \\
  \texttt{\small rajarshi.das@alphainfra.ai}
  \And
  Aditya Dubey \\
  \textbf{\small AlphaInfra} \\
  \texttt{\small aditya.d@alphainfra.ai}
  \AND
  Manminder Tomar \\
  \textbf{\small AlphaInfra} \\
  \texttt{\small manminder.t@alphainfra.ai}
  \And
  Sreekar Reddy \\
  \textbf{\small AlphaInfra} \\
  \texttt{\small sreekar.r@alphainfra.ai}
  \And
  Samta Sharma \\
  \textbf{\small AlphaInfra} \\
  \texttt{\small samta.s@alphainfra.ai}
  \AND
  Satyapragnya Kar\footnotemark[1] \\
  \textbf{\small New York University} \\
  \texttt{\small sk13001@nyu.edu}
  \And
  PR Panigrahi \\
  \textbf{\small Envana Software Solutions} \\
  \texttt{\small pr.panigrahi@envana.com}
}

\begin{document}
\maketitle
\setcounter{footnote}{0}

\begin{abstract}
An AI data center is committed years before it draws a watt: a few discrete
choices --- grid or on-site gas, evaporative or dry cooling, which county to
file in --- fix its capital, water draw, and carbon path, and LLMs are now
asked to help screen them with no test of whether they track what each
commitment rules out. We introduce DCBench-440, with two arms reported
separately: \nGenQ{} generative questions over STRIPS worlds for five US
deal archetypes, with search-verified gold answers and no tested shortcut
above \ProbeGenMax\% on any question type, and \nMcqQ{} multiple-choice
questions grounded in quoted site records, emission equations, and stated
scenarios. Across open-weight deployments, accuracy spans
\GenOpenMin--\GenOpenMax\% (generative) and \McqOpenMin--\McqOpenMax\%
(multiple choice). One skill fails everywhere: asked to list every action
available at a decision gate, no model exceeds \AppStrictMax\% exact match.
The strongest list \AppRecChgMin--\AppRecChgMax\% of the actions that would
change the project's state but omit most permitted actions that change
nothing; with those excused, the best exact match is still \AppExclMax\%.
Code, data, and other materials are available upon request.
\end{abstract}

\section{Introduction}

The AI data center buildout is one of the largest capital programs in
history, roughly \$7\,trillion by 2030, with power and water demand climbing
alongside \cite{mckinsey2025,vannieuwerburgh2026,cbre2024,iea2025,gs2024,epri2024}. The hard
question is the order of commitments. Take a developer we work with in the
US:\footnote{Site names and counties are withheld. Figures are from the
developer's own financing materials.} one site plans 550\,MW of on-site gas
and holds every permit but the air permit; a second needs 750\,MW from the
grid and is one of seven waiting on the utility; a third is bare land with a
gas pipeline across it. On-site gas skips a grid queue where capacity now
waits three to seven years \cite{lbnl2026}, but takes on turbines, fuel
supply, and emissions permits. Developers are taking that trade: 61\% of
those surveyed name on-site power their primary strategy when the grid is
the bottleneck \cite{bloom2026}, and 124\,GW of on-site gas has been
announced for US data centers \cite{bnef2026}. The choices that fix a site's water and
carbon path --- evaporative or dry cooling, on-site gas or grid --- also fix
its capital cost and date of first power, so at each gate the question is
what a commitment rules out.

Language models are entering this work as screening and diligence aids, and
LLM analysis of corporate, policy, and media sources already feeds
projections of data-center power demand \cite{chen2026aidc}. A fluent
answer is not a plan. A wrong number can be checked; an option the
model never mentions --- a permit that must come first, a route that a
choice forecloses --- cannot.

DCBench-440 measures that second kind of failure. It contributes:
\begin{itemize}
  \item a generative arm applying ACPBench Hard's eight question types
  \cite{acpbenchhard} to five US deal archetypes, selected against shortcuts
  (Section~\ref{sec:gen});
  \item an independent multiple-choice arm (Section~\ref{sec:mcq});
  \item an evaluation of \nOpenDeploy{} deployments of \nOpenWeights{}
  open-weight models with intervals, paired tests, settings, and cost
  (Section~\ref{sec:results}, Appendix~\ref{app:extended}); and
  \item a diagnosis of the failure every model shares: exact enumeration of
  the actions available at a gate, that is, listing every step the project
  could take next.
\end{itemize}

\section{Background and related work}
\label{sec:background}

Each generative-arm problem is a STRIPS task \cite{strips} written in PDDL
\cite{pddl}, $\Pi = \langle F, A, s_0, s^\star \rangle$, with boolean facts
$F$, states $s \subseteq F$, an initial state $s_0$, and a goal $s^\star$. An
action $a = \langle pre(a), add(a), del(a)\rangle$ applies in $s$ when
$pre(a) \subseteq s$ and yields $(s \setminus del(a)) \cup add(a)$; we call
it a \emph{no-op} in $s$ when that result is $s$ itself. In plain terms, a
state is the set of facts that are currently true for a project (for
example, whether a permit has been obtained); an action can be taken only
when its preconditions hold, and it then makes some facts true and others
false. A no-op is an action that is allowed but changes nothing, because its
effects already hold. As in ACPBench
\cite{acpbench,acpbenchhard}, the model must produce an answer rather than
choose one: each prompt gives the world in plain text, the state or plan in
question, and two worked examples from a held-out tutorial domain, and a
lenient parser and a symbolic validator grade the reply (unparseable counts
as wrong).

\paragraph{Related work.} PlanBench \cite{planbench} and critical studies of
LLM planning \cite{valmeekam2023planning} score whole plans on puzzle and
logistics domains. ACPBench \cite{acpbench} asks atomic planning questions as
yes/no and multiple choice; ACPBench Hard \cite{acpbenchhard} asks
open-ended versions of the same tasks, and there GPT-4o, which answers 97\%
of yes/no and 89\% of multiple-choice applicability questions correctly,
solves 25\% when it must produce the applicable set. Hybrid systems hand
planning to a symbolic planner \cite{liu2023llmp} or check LLM proposals
with external verifiers \cite{kambhampati2024modulo}; we measure the
language model alone, as screening assistants are used today. Generation and
verification can come apart in language models
\cite{west2024paradox,li2024gvconsistency}, and self-verification of plans
is unreliable \cite{stechly2024selfverify}. Multiple-choice scores can
reflect option artifacts \cite{balepur2024artifacts} and position bias
\cite{zheng2024mcq}; our gold letters are balanced across positions, and we
test option length directly.

\paragraph{Data-center planning and AI's footprint.} Data-center siting has been posed as
county-level cost optimization \cite{vabson2026siting}, geospatial
suitability analysis \cite{arzumanyan2025texas}, and what-if simulation of
sustainable deployment \cite{tsiopani2026infactplanner}; these compute a
decision rather than test a model's reasoning about one, and a review of 189
ML-assisted site-selection studies lists the absence of benchmark datasets
among the field's main challenges \cite{upadhyay2026review}. Work on AI's environmental cost measures what
models themselves consume \cite{patterson2021,unu2026}; question sets on
data-center economics test recall or use a model as judge.

\section{Two arms}

\subsection{The generative arm: five deal archetypes}
\label{sec:gen}

Each domain, specified with an industry team that develops and underwrites
such projects, models one way US
data centers get built: \code{btm\_ccgt}, a campus on its own gas turbines
gated by air permit, gas lateral, and fuel contract; \code{grid\_queue},
utility interconnection from screening to security; \code{retrofit},
densifying a live site under a feeder budget; \code{gulf\_water}, a coastal
site whose one-shot cooling fork sets its water draw; and
\code{permian\_gas}, gensets at well pads (Appendix Table~\ref{tab:domains}).
Environmental, regulatory, and financial content enters as structure, not
arithmetic: a \emph{static} fact, which no action adds or deletes --- a gas
line at the fence, water-district capacity, air-quality nonattainment ---
closes off part of the action space for good and holds in some instances but
not others, giving reachability and landmark \cite{landmarks} questions real
answers. Technology choices are one-shot forks, and gated alternative routes
(modular units, bridge fuel, rental chillers, prefabricated blocks) vary
landmark structure within a domain. These devices mirror how an active
origination pipeline screens and stages sites (Appendix~\ref{app:screening});
project metadata is synthetic. Gold answers come from breadth-first search;
a question is dropped, not guessed, when search hits its node cap, so worlds
stay small. Each domain
contributes five questions per type (Table~\ref{tab:questions}), scored 0/1,
and a symbolic oracle scoring 100\% is a standing test of the grader.

\begin{table}[t]
  \caption{The eight generative question types (from ACPBench Hard);
  $h^\star$ is the optimal cost.}
  \label{tab:questions}
  \centering
  \small
  \begin{tabularx}{\linewidth}{@{}l >{\raggedright\arraybackslash}X
      >{\raggedright\arraybackslash}X >{\raggedright\arraybackslash}p{2.45cm}@{}}
    \toprule
    Type & The question & Gold answer & Graded by \\
    \midrule
    \textsc{App}    & every action applicable right now & the full applicable set, no-ops included & exact set match \\
    \textsc{Prog}   & what one action adds and deletes & $add(a)$ and $del(a)$ & exact set match \\
    \textsc{Reach}  & a fact that can never hold & facts in no reachable state & membership \\
    \textsc{AReach} & an action that can never apply & actions applicable in no reachable state & membership \\
    \textsc{Val}    & the first step of a plan that breaks & index of the first inapplicable action & exact index \\
    \textsc{Just}   & a plan step that buys nothing & actions whose removal leaves a valid plan & simulate the shortened plan \\
    \textsc{Land}   & a milestone every plan must pass & landmark facts of the instance & membership \\
    \textsc{NextA}  & a step that provably shortens the path & actions that reduce $h^\star$ & membership \\
    \bottomrule
  \end{tabularx}
\end{table}

\begin{figure}[t]
  \centering
  \begin{tikzpicture}[
      every node/.append style={execute at begin node={\hyphenpenalty=10000}},
      font=\figtxt,
      pill/.style={rounded corners=2pt, draw=black!65, fill=black!3,
                   font=\figmono, inner xsep=2pt, inner ysep=2pt,
                   text width=2.30cm, align=center, minimum height=0.82cm},
      dead/.style={pill, draw=black!40, dashed, fill=white, text=black!55},
      goal/.style={pill, draw=black!75, double, fill=black!8},
      band/.style={draw=black!45, fill=black!5, rounded corners=2pt,
                   inner sep=3.2pt, font=\figtxt},
      grp/.style={draw=black!55, rounded corners=2pt, fill=white,
                  inner sep=3.4pt, align=left, font=\figtxt},
      go/.style={-{Stealth[length=1.6mm]}, draw=black!70},
      hint/.style={-{Stealth[length=1.3mm]}, draw=black!45},
      num/.style={font=\fignum, text=black!60, inner sep=1pt},
    ]

    \node[band, text width=13.4cm, align=center] (statics) at (6.85,3.30)
      {\textbf{static facts} (no action adds or deletes them)\\
       \texttt{(water-district-capacity)} present\ \ \checkmark\quad$\vert$\quad
       \texttt{(dry-cooling-viable)} absent\ \ $\times$};

    \node[pill] (a1) at (1.32,2.15) {(secure-site)};
    \node[pill] (a2) at (4.08,2.15) {(mobilize-crew)};
    \node[pill] (a3) at (6.85,2.15) {(select-evap-\\cooling)};
    \node[pill] (a4) at (9.61,2.15) {(contract-water)};
    \node[pill] (a5) at (12.38,2.15) {(build-cooling-\\evap)};
    \foreach \i/\j in {a1/a2, a2/a3, a3/a4, a4/a5}
      \draw[go] (\i) -- (\j);
    \foreach \n/\i in {1/a1, 2/a2, 3/a3, 4/a4, 5/a5}
      \node[num, above=0pt of \i.north] {\n};

    \node[pill] (a6) at (1.32,0.86) {(file-tceq)};
    \node[pill] (a7) at (4.08,0.86) {(build-block b1)};
    \node[pill] (a8) at (6.85,0.86) {(obtain-tceq)};
    \node[pill] (a9) at (9.61,0.86) {(energize-\\block b1)};
    \node[goal] (gl) at (12.38,0.86) {goal reached:\\(energized b1)};
    \foreach \i/\j in {a6/a7, a7/a8, a8/a9, a9/gl}
      \draw[go] (\i) -- (\j);
    \foreach \n/\i in {6/a6, 7/a7, 8/a8, 9/a9}
      \node[num, above=0pt of \i.north] {\n};

    \node[dead] (dd) at (6.85,-0.22) {(select-dry-\\cooling)};
    \draw[hint, dashed] (a3.south) -- (6.85,1.64) -- (8.23,1.64) --
                        (8.23,0.33) -- (6.85,0.33) -- (dd.north);
    \node[text=black!55, right=1.5mm of dd, text width=5.2cm,
          align=left] (ddnote)
      {never applicable in any reachable state, because
       \texttt{(dry-cooling-viable)} is absent};
    \node[text=black!55, left=1.5mm of dd, text width=4.6cm,
          align=right]
      {step 3 is a \textbf{one-shot fork}: it consumes\\
       \texttt{(cooling-undecided)} for good};

    \node (hdr) at (6.85,-1.02)
      {\textbf{the eight question types, every one asked about the world above}};
    \node[grp, text width=4.05cm, anchor=north west] (g1) at (0.10,-1.24)
      {\textbf{asked of one action}\\
       \textsc{App} every action available now\\
       \textsc{Prog} what one action changes};
    \node[grp, text width=4.05cm, anchor=north] (g2) at (6.85,-1.24)
      {\textbf{asked of the state space}\\
       \textsc{Reach} a fact that can never hold\\
       \textsc{AReach} work that can never happen};
    \node[grp, text width=4.05cm, anchor=north east] (g3) at (13.60,-1.24)
      {\textbf{asked of a plan}\\
       \textsc{Val} first step that breaks\\
       \textsc{Just} a step that buys nothing\\
       \textsc{Land} a milestone every plan needs\\
       \textsc{NextA} a step that provably shortens};

  \end{tikzpicture}
  \caption{A one-block world of the \code{gulf\_water} domain and the eight
  generative-arm questions asked about it (benchmark instances have two or
  three blocks, and some add rental-chiller or prefabricated-block routes).
  The nine numbered pills are an optimal plan, read left to right and
  wrapping to the second row. Because the static fact
  \code{(dry-cooling-viable)} is absent, the dry-cooling branch is dead
  forever: only two actions apply at the start,
  \code{(cooling-dry-selected)} can never hold, and \code{(water-contracted)}
  is a landmark in \emph{this} instance though it would not be if dry cooling
  were viable. The statics, not the wording, decide the answer.}
  \label{fig:schematic}
\end{figure}
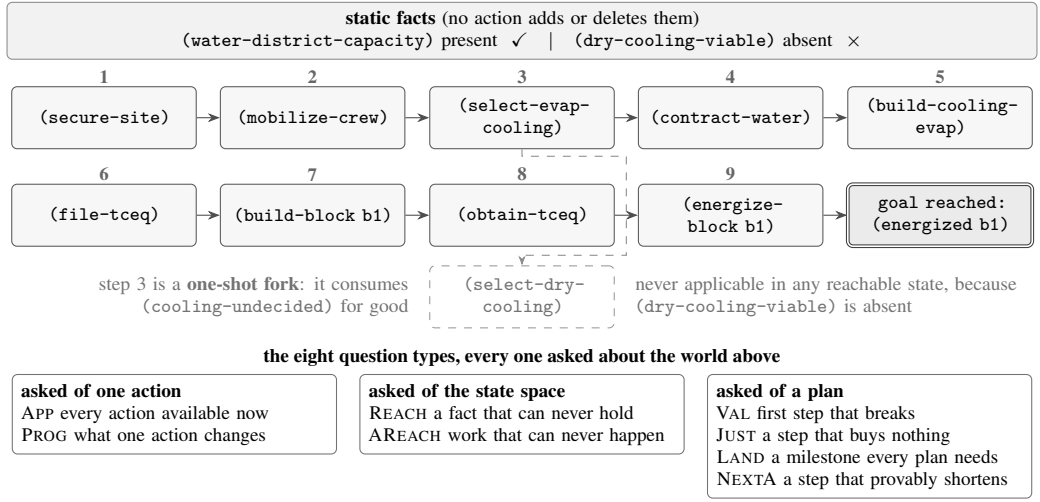

\paragraph{A worked example.} Figure~\ref{fig:schematic} shows a small
world of the \code{gulf\_water} domain, an optimal plan through it, and the
eight generative-arm questions asked of it. The point of the example is the
static fact \code{(dry-cooling-viable)}: because it is absent in this
instance, the dry-cooling branch is dead from the start, and that single fact
decides the gold answer to the applicability, reachability, and landmark
questions.

\paragraph{Selected against shortcuts.} Planning questions invite
shortcuts: a question can sometimes be answered from surface patterns in the
prompt, without simulating the world. We therefore use worlds large enough
to resist such patterns (\WorldStates{} reachable states on average) yet
small enough for exact search, estimate for every candidate
question the chance that the best cheap strategy is credited, and keep the
least guessable. Scored through the same parser and validators, no
world-model-free strategy we tested exceeds \ProbeGenMax\% on any question
type (Appendix Table~\ref{tab:gen_task_full}).

\subsection{The multiple-choice arm: source-grounded questions}
\label{sec:mcq}

The second arm comes from two separately authored 120-question banks over
six content buckets: environmental impact, power infrastructure, AI
workload, financial and siting decisions, cross-domain, and permitting. Each
question carries one of eight skill labels; four share a name with a
generative type, but none is the same task: multiple-choice applicability
asks which stated option complies with a stated limit, not which actions
are available. A question's context block holds every fact it needs: the
first bank quotes a 1{,}677-row database of real US sites, published
emission equations, vendor calculators, and industry reports; the second states
self-contained scenario assumptions in five buckets and quotes the site
database in the sixth. Models answer zero-shot with one letter, graded by
exact match. Picking the longest option with the question unread scores
\McqLongest\% (chance: 25\%), so we also report the \McqHardN{} questions
where the longest option is wrong. As the arms differ in content, skill
definitions, prompting, and grading, we report them separately and read any
gap between them as descriptive, not as a format effect.

\section{Results}
\label{sec:results}

We evaluate \nOpenDeploy{} deployments of \nOpenWeights{} open-weight models
on Groq and OpenRouter at temperature 0, without fine-tuning or tools
(settings, tokens, and cost in Appendix Table~\ref{tab:settings}; OpenRouter
charged \$\CostOpenRouter{} for both arms). Serving GPT-OSS through either
provider moves generative accuracy by \ProvGapMin--\ProvGapMax{} points
($p\ge\ProvGapPMin$); Qwen3.8-27B differs by \QwenGap{} points ($p=\QwenGapP$),
but Groq runs it with reasoning off and OpenRouter with it on; Qwen3.6-27B,
with the same mismatch, is held out pending a matched re-run.
Groq Compound and Compound-mini are hosted systems, not open-weight models,
and ran with server-side tools on --- Compound's traces show tool calls,
chiefly Python, in \ToolCompoundGen/200 generative and \ToolCompoundMcq/240
multiple-choice answers (Compound-mini: \ToolCompoundMiniGen{} and
\ToolCompoundMiniMcq) --- so we report them apart. Multiple-choice runs exist for \nMcqOpenRun{} of
the \nOpenDeploy{} open-weight deployments.

\begin{figure}[t]
  \centering
  \includegraphics[width=\linewidth]{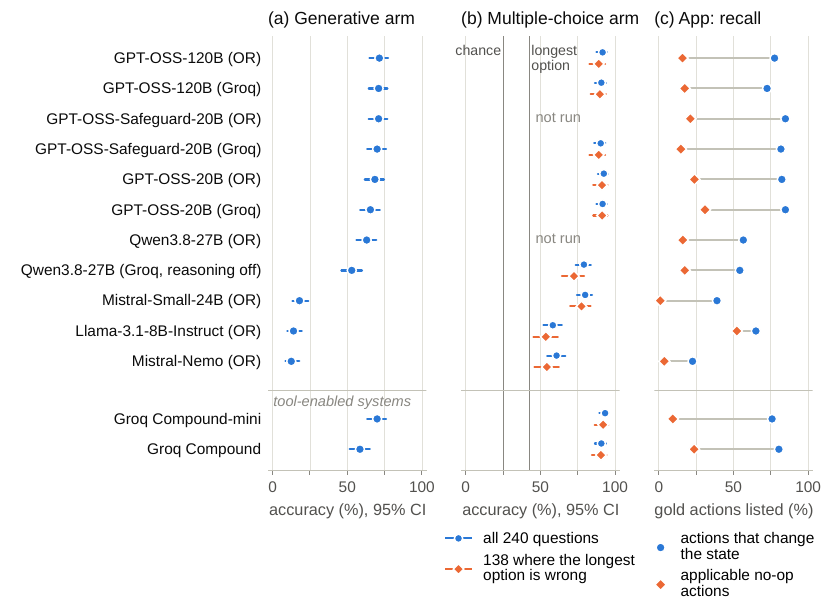}
  \caption{Per-arm results. Each row is one deployment. In (a) and (b), a
  marker is its accuracy and the bar through it the 95\% Wilson interval.
  (a)~Generative arm.
  (b)~Multiple-choice arm: all 240 questions (circles) and the \McqHardN{}
  where the longest option is wrong (diamonds); lines mark chance and the
  longest option. (c)~Generative \textsc{App}: share of gold actions
  listed that change the state (circles) or are no-ops (diamonds). Below the
  line: Groq Compound systems, tools on. OR: OpenRouter.}
  \label{fig:arms}
\end{figure}

\paragraph{Per-arm accuracy.} Each deployment gave one scored answer per
question, so we report accuracy with 95\% Wilson intervals, which stay
reliable near 0 and 100\%, and compare deployments with exact McNemar tests,
which use only the questions on which exactly one of the two is correct. The 95\% intervals span
$\pm$\GenHalfMin--\GenHalfMax{} points on the generative arm and
$\pm$\McqHalfMin--\McqHalfMax{} on multiple choice (Figure~\ref{fig:arms}a--b,
Appendix Table~\ref{tab:arms}), and neither arm has a meaningful leader: the
\GenTiedN{} GPT-OSS deployments (\GenTiedMin--\GenTiedMax\%) are not
significantly different from the best on the generative arm (exact McNemar,
$p\ge0.05$), nor are the \McqTiedN{} with multiple-choice runs
(\McqTiedMin--\McqTiedMax\%) on the other. The three weakest models nearly
fail the generative arm (\WeakGenMin--\WeakGenMax\%) while scoring
\WeakMcqMin--\WeakMcqMax\% on multiple choice. Strong multiple-choice scores
do not rest on option length: where the longest option is wrong, the
GPT-OSS deployments score \McqHardStrongMin--\McqHardStrongMax\%, while weaker
ones lose \McqHardDropMin--\McqHardDropMax{} points.

\paragraph{One shared failure: listing what can be done next.} Asked for
every action applicable in a state, no deployment exceeds \AppStrictMax\%
exact match (a reply counts only if the listed set equals the gold set), and \AppZeroOpen{} of the \nOpenDeploy{} open-weight
deployments score 0.0\% (Appendix Table~\ref{tab:app_full}). Gold sets
average \AppGoldMean{} actions, and \AppNoopShare\% of their members are
no-ops, which the prompt explicitly says to include. Figure~\ref{fig:arms}c
separates the two kinds: the six GPT-OSS deployments list
\AppRecChgMin--\AppRecChgMax\% of the gold actions that would change the
state but only \AppRecNoopMin--\AppRecNoopMax\% of the no-ops, which make up
\AppOmitNoopMin--\AppOmitNoopMax\% of everything they omit, at precision
\AppPrecMin--\AppPrecMax. The convention is not the whole failure: they
still miss \AppMissChgMin--\AppMissChgMax\% of the state-changing actions,
and scored without no-ops on either side, the best exact match is
\AppExclMax\%. Errors are rarely formatting: at most \FmtStrongMax{} of 200 answers per stronger
deployment are unparseable, ungrounded, or truncated (up to \FmtWeakMax{} for
the weakest, mostly naming facts or actions that do not exist; Appendix
Table~\ref{tab:outcomes}).

\paragraph{Partial credit needs a baseline.} Listing every grounded action
without reading the state scores F1 \ProbeAppF{} and Jaccard \ProbeAppJ,
above every model (best \AppFoneMax{} and \AppJaccardMax), and clears
Jaccard 0.5 on \ProbeAppJfifty\% of questions; only exact match and high
thresholds reject it (Jaccard $\ge 0.8$: 0\% against \AppJeightyMax\% for
the best model). We therefore report exact match, precision (the share of
listed actions that are applicable), and recall (the share of applicable
actions that are listed).

\paragraph{What the two arms do and do not show.} Every deployment scores
higher on multiple choice, but the arms differ in content and grading, and
multiple-choice applicability (\McqAppMin--\McqAppMax\%) checks compliance,
not enumeration: the gap is not a format effect. It does show that high
accuracy on source-grounded multiple choice does not imply the ability to
enumerate a gate (the open-weight deployments above 90\% there score
\AppStrictStrongMax\% on exact enumeration). Posing the same gate in each
format is next.

\section{Limitations, impact, and pathway to climate impact}
\label{sec:limits}

\paragraph{Limitations.}
\begin{itemize}
  \item Worlds are small so that search stays exact, their metadata is
  synthetic, and five archetypes are a narrow base (a sixth is paused for
  rework).
  \item Each deployment answered once, and temperature 0 was not
  deterministic on these APIs.
  \item We test no frontier closed models, have no human baseline, and lack
  multiple-choice runs for two deployments.
\end{itemize}

\paragraph{Pathway to climate impact.} DCBench-440 scores no emissions or water
quantities, and we do not claim that better planning lowers emissions by
itself. A site's water draw and emissions are set by a few early,
irreversible choices --- the cooling fork, the power pathway, the permits
that gate them --- now often screened with LLM help. On the power pathway alone,
at least 74 gas plants proposed to serve US data centers directly would emit
nearly 662 million tons of greenhouse gases a year \cite{eip2026}. An assistant that omits
available options shrinks the set a developer weighs; omission is the
failure we find in every model, and measuring it is a precondition for
trusting such tools. Next, we will attach water and emissions quantities to
each world's branches so an omission can be weighted by what it forecloses.

\paragraph{Dual use and value beyond data centers.} The benchmark is technology-neutral --- a better planner can speed a
gas-fired build as readily as a dry-cooled or grid-tied one --- and a fixed
test set can be optimized against, so worlds are regenerated. Beyond data
centers, the diagnosis applies to any agent that must enumerate its options.

\paragraph{Availability.} Code, data, and other materials are available from
the authors upon request.

{\small\raggedright
\bibliographystyle{plainnat}
\bibliography{references}
}

\clearpage
\appendix

\section{Screening funnel and stage-gate schematic}
\label{app:screening}

Section~\ref{sec:gen} notes that the generative arm's static facts, one-shot
forks, and landmarks are not chosen for formal convenience: they mirror how
an active U.S.\ origination pipeline actually screens and sequences a site.
We reviewed that pipeline's screening and staging materials in confidence,
and we do not reproduce them here; this appendix instead schematizes, at the
same level of generality already stated in Section~\ref{sec:gen}, the
two-stage process that motivated our domain design. No confidential
thresholds, company names, or deal-specific figures appear below or anywhere
else in this paper.

Figure~\ref{fig:screening_funnel} shows the first stage: a candidate site is
screened against a small set of threshold tests, each of which is a static
fact in our domains because it either holds or does not and, when it fails,
removes a route rather than merely lowering a score. Figure~\ref{fig:stage_gate}
shows the second stage: a site that clears the screen still moves through an
ordered sequence of commitments, and the financial investment decision is
the one-shot fork in this sequence --- capital is released to retire a
single named risk, and that choice is what our domains encode as a
consumable resource being spent for good rather than as a number going up.

\begin{figure}[h]
  \centering
  \begin{tikzpicture}[
      every node/.append style={execute at begin node={\hyphenpenalty=10000}},
      font=\footnotesize,
      test/.style={rounded corners=2pt, draw=black!65, fill=black!3,
                   text width=5.2cm, align=center, inner sep=4pt},
      pass/.style={rounded corners=2pt, draw=black!75, double, fill=black!8,
                   text width=3.2cm, align=center, inner sep=5pt},
      fail/.style={rounded corners=2pt, draw=black!40, dashed, fill=white,
                   text width=4.6cm, align=center, inner sep=4pt, text=black!55},
      arr/.style={-{Stealth[length=1.4mm]}, draw=black!70},
      thin_arr/.style={draw=black!55, thin},
    ]
    \node[test] (t1) at (0,4.8) {Contiguous acreage under single-owner control};
    \node[test] (t2) at (0,3.6) {Heavy-industrial zoning already granted};
    \node[test] (t3) at (0,2.4) {High-voltage substation on site or within a few miles};
    \node[test] (t4) at (0,1.2) {Two or more gas pipelines at the fence line};
    \node[test] (t5) at (0,0)   {Permitted water rights};

    \coordinate (bus) at (5.0,2.4);
    \foreach \i in {t1,t2,t3,t4,t5}
      \draw[thin_arr] (\i.east) -| (bus);

    \node[pass] (pass) at (8.0,2.4) {Candidate site\\advances to stage gates\\(Fig.~\ref{fig:stage_gate})};
    \draw[arr] (bus) -- (pass.west);

    \node[fail] (fail) at (8.4,-0.75) {Any one test fails: that development route is closed for good, not merely deprioritized};
    \draw[arr, dashed] (bus) |- (fail.west);

  \end{tikzpicture}
  \caption{The screening stage. Each threshold test is a static fact: it is
  true or false for a given site and no later action changes it. Passing
  requires every test to hold at once, so failing even one removes a route
  permanently --- the same behavior our \code{static} predicates encode in
  the generative-arm domains (Section~\ref{sec:gen}).}
  \label{fig:screening_funnel}
\end{figure}
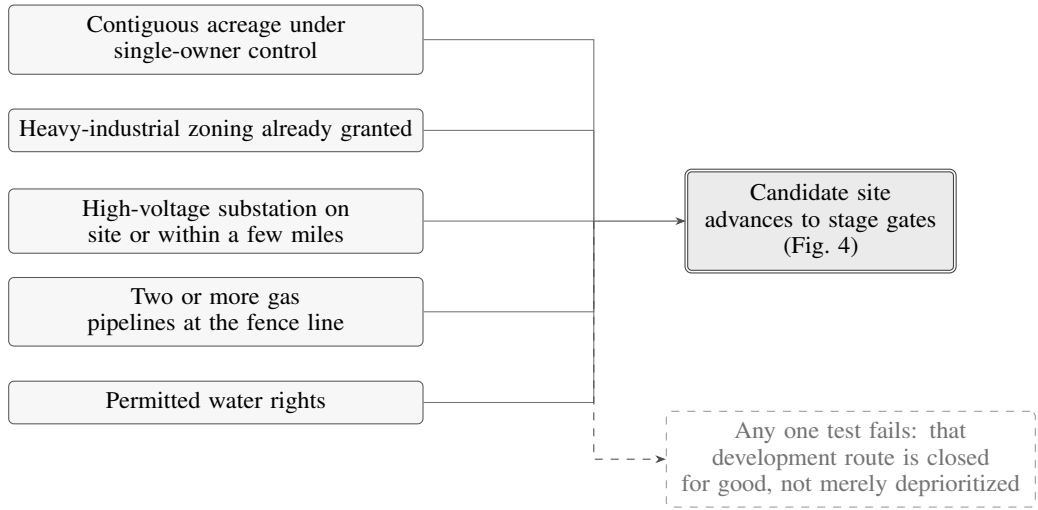

\begin{figure}[h]
  \centering
  \begin{tikzpicture}[
      every node/.append style={execute at begin node={\hyphenpenalty=10000}},
      font=\footnotesize,
      gate/.style={rounded corners=2pt, draw=black!65, fill=black!3,
                   text width=2.75cm, align=center, inner sep=4pt},
      fid/.style={rounded corners=2pt, draw=black!75, double, fill=black!8,
                   text width=3.0cm, align=center, inner sep=5pt},
      arr/.style={-{Stealth[length=1.4mm]}, draw=black!70},
      note/.style={font=\scriptsize, text=black!55, align=center, text width=3.5cm},
    ]
    \node[gate] (g1) at (0,0)     {Site control (screen passed)};
    \node[gate] (g2) at (3.45,0)  {Development capital: permitting and equipment procurement};
    \node[fid]  (g3) at (7.0,0)   {Financial investment decision: capital released to retire one named risk};
    \node[gate] (g4) at (10.55,0) {Build and energize};

    \draw[arr] (g1) -- (g2);
    \draw[arr] (g2) -- (g3);
    \draw[arr] (g3) -- (g4);

    \node[note] at (0,-1.45)   {an early commitment here fixes what remains feasible later};
    \node[note] at (7.0,-1.45) {a one-shot fork: this commitment cannot be undone};
  \end{tikzpicture}
  \caption{The staging stage. A site that clears Figure~\ref{fig:screening_funnel}
  still moves through an ordered chain of commitments rather than a single
  go/no-go decision. The financial investment decision is a one-shot fork in
  our domains: the resource it spends (capital committed against a named
  risk) cannot be recovered by a later action, which is why an early stage
  gate can foreclose options several stages downstream.}
  \label{fig:stage_gate}
\end{figure}
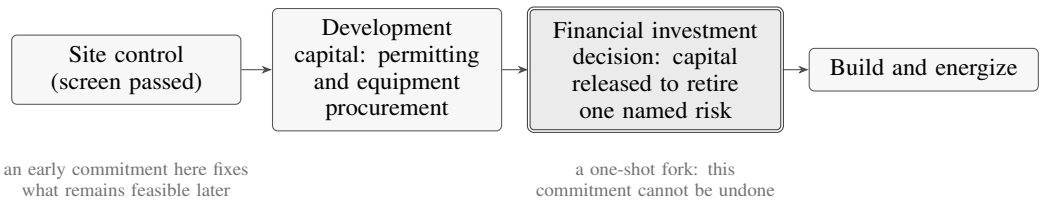

\clearpage
\section{Extended results}
\label{app:extended}

Every table in this appendix is generated by
a script from the stored model
responses, rescored through the benchmark's own parser and validators and
checked against the committed score files; no number is typed by hand.
Table~\ref{tab:arms} gives each arm with intervals and paired tests;
Tables~\ref{tab:gen_task_full} and~\ref{tab:gen_domain_full} break the
generative arm down by question type and domain, with the shortcut
baselines; Tables~\ref{tab:mcq_task_full} and~\ref{tab:mcq_domain_full} do
the same for the multiple-choice arm; Table~\ref{tab:app_full} details
applicability; Table~\ref{tab:outcomes} classifies every generative answer;
Table~\ref{tab:settings} gives inference settings, tokens, cost, and tool
use; and Table~\ref{tab:domains} gives domain sizes.

\begin{table}[!htbp]
  \caption{Accuracy (\%) on each arm with 95\% Wilson intervals. $p$: exact McNemar test against the best open-weight deployment on that arm (paired by question). OR: OpenRouter. LO-wrong: the 138 multiple-choice questions on which the longest option is not the answer. Bank 1 and Bank 2 are the two independently authored 120-question banks. The arms differ in content, skills, prompting, and grading, so the two columns are not a format comparison. ``Not run'': no multiple-choice run exists for that deployment.}
  \label{tab:arms}
  \centering
  \scriptsize
  \setlength{\tabcolsep}{3.5pt}
  \begin{tabular}{@{}lrcr rcrr rr@{}}
    \toprule
    Deployment & \multicolumn{3}{c}{Generative (200)} & \multicolumn{3}{c}{Multiple choice (240)} & LO-wrong & Bank 1 & Bank 2 \\
    \cmidrule(lr){2-4}\cmidrule(lr){5-7}
    & Acc. & 95\% CI & $p$ & Acc. & 95\% CI & $p$ & (138) & (120) & (120) \\
    \midrule
    Symbolic oracle & 100.0 & [98.1, 100.0] & -- & -- & -- & -- & -- & -- & -- \\
    GPT-OSS-120B (OR) & 71.5 & [64.9, 77.3] & -- & 91.7 & [87.5, 94.5] & 0.754 & 89.1 & 95.0 & 88.3 \\
    GPT-OSS-120B (Groq) & 71.0 & [64.4, 76.8] & 1.000 & 90.8 & [86.5, 93.9] & 0.344 & 89.9 & 93.3 & 88.3 \\
    GPT-OSS-Safeguard-20B (OR) & 71.0 & [64.4, 76.8] & 1.000 & \multicolumn{6}{c}{not run} \\
    GPT-OSS-Safeguard-20B (Groq) & 70.0 & [63.3, 75.9] & 0.749 & 90.4 & [86.0, 93.5] & 0.180 & 89.1 & 91.7 & 89.2 \\
    GPT-OSS-20B (OR) & 68.5 & [61.8, 74.5] & 0.345 & 92.5 & [88.5, 95.2] & -- & 91.3 & 93.3 & 91.7 \\
    GPT-OSS-20B (Groq) & 65.5 & [58.7, 71.7] & 0.096 & 91.7 & [87.5, 94.5] & 0.688 & 91.3 & 93.3 & 90.0 \\
    Qwen3.8-27B (OR) & 63.0 & [56.1, 69.4] & 0.019 & \multicolumn{6}{c}{not run} \\
    Qwen3.8-27B (Groq) & 53.0 & [46.1, 59.8] & $<$0.001 & 79.2 & [73.6, 83.8] & $<$0.001 & 72.5 & 80.8 & 77.5 \\
    Mistral-Small-24B (OR) & 18.0 & [13.3, 23.9] & $<$0.001 & 80.0 & [74.5, 84.6] & $<$0.001 & 77.5 & 80.8 & 79.2 \\
    Llama-3.1-8B-Instruct (OR) & 14.0 & [9.9, 19.5] & $<$0.001 & 58.3 & [52.0, 64.4] & $<$0.001 & 53.6 & 50.0 & 66.7 \\
    Mistral-Nemo (OR) & 12.5 & [8.6, 17.8] & $<$0.001 & 60.8 & [54.5, 66.8] & $<$0.001 & 54.3 & 62.5 & 59.2 \\
    \midrule
    \multicolumn{10}{@{}l}{\emph{Tool-enabled systems (server-side tools on; not open-weight)}}\\
    Groq Compound-mini & 70.0 & [63.3, 75.9] & -- & 93.3 & [89.4, 95.9] & -- & 92.0 & 95.8 & 90.8 \\
    Groq Compound & 58.5 & [51.6, 65.1] & -- & 90.8 & [86.5, 93.9] & -- & 90.6 & 93.3 & 88.3 \\
    \midrule
    Chance & -- & -- & -- & 25.0 & -- & -- & -- & -- & -- \\
    Longest option, question unread & -- & -- & -- & 42.5 & -- & -- & 0.0 & 34.2 & 50.8 \\
    \bottomrule
  \end{tabular}
\end{table}

\begin{table}[!htbp]
  \caption{Generative-arm accuracy (\%) by question type, 25 questions per type, so each cell carries a 95\% interval of up to $\pm$19 points. \emph{Best shortcut}: the highest accuracy any world-model-free strategy reaches through the same parser and validators (for \textsc{App}: listing every grounded action). \emph{Shortcut bound}: the generator's analytic upper bound on guessing, used to select the least guessable questions.}
  \label{tab:gen_task_full}
  \centering
  \scriptsize
  \setlength{\tabcolsep}{3.5pt}
  \begin{tabular}{@{}lrrrrrrrr@{}}
    \toprule
    Deployment & \textsc{App} & \textsc{Prog} & \textsc{Reach} & \textsc{AReach} & \textsc{Val} & \textsc{Just} & \textsc{Land} & \textsc{NextA} \\
    \midrule
    Symbolic oracle & 100.0 & 100.0 & 100.0 & 100.0 & 100.0 & 100.0 & 100.0 & 100.0 \\
    GPT-OSS-120B (OR) & 0.0 & 88.0 & 96.0 & 48.0 & 76.0 & 100.0 & 80.0 & 84.0 \\
    GPT-OSS-120B (Groq) & 0.0 & 92.0 & 92.0 & 36.0 & 88.0 & 100.0 & 84.0 & 76.0 \\
    GPT-OSS-Safeguard-20B (OR) & 0.0 & 96.0 & 96.0 & 48.0 & 100.0 & 100.0 & 60.0 & 68.0 \\
    GPT-OSS-Safeguard-20B (Groq) & 0.0 & 96.0 & 96.0 & 52.0 & 92.0 & 88.0 & 68.0 & 68.0 \\
    GPT-OSS-20B (OR) & 0.0 & 84.0 & 96.0 & 36.0 & 96.0 & 100.0 & 60.0 & 76.0 \\
    GPT-OSS-20B (Groq) & 0.0 & 92.0 & 88.0 & 20.0 & 92.0 & 88.0 & 72.0 & 72.0 \\
    Qwen3.8-27B (OR) & 4.0 & 72.0 & 92.0 & 64.0 & 48.0 & 100.0 & 48.0 & 76.0 \\
    Qwen3.8-27B (Groq) & 0.0 & 80.0 & 72.0 & 44.0 & 80.0 & 64.0 & 16.0 & 68.0 \\
    Mistral-Small-24B (OR) & 0.0 & 20.0 & 24.0 & 48.0 & 8.0 & 0.0 & 0.0 & 44.0 \\
    Llama-3.1-8B-Instruct (OR) & 0.0 & 8.0 & 44.0 & 28.0 & 20.0 & 4.0 & 0.0 & 8.0 \\
    Mistral-Nemo (OR) & 0.0 & 28.0 & 8.0 & 8.0 & 20.0 & 24.0 & 0.0 & 12.0 \\
    \midrule
    \multicolumn{9}{@{}l}{\emph{Tool-enabled systems (server-side tools on; not open-weight)}}\\
    Groq Compound-mini & 0.0 & 92.0 & 100.0 & 32.0 & 80.0 & 96.0 & 80.0 & 80.0 \\
    Groq Compound & 4.0 & 60.0 & 84.0 & 28.0 & 88.0 & 80.0 & 68.0 & 56.0 \\
    \midrule
    Best shortcut (measured) & 0.0 & 0.0 & 8.0 & 0.0 & 12.0 & 4.0 & 12.0 & 0.0 \\
    Shortcut bound ($\overline{\mathit{guess\_p}}$) & 0.0 & 0.0 & 29.2 & 18.3 & 8.4 & 4.9 & 21.8 & 11.1 \\
    \bottomrule
  \end{tabular}
\end{table}

\begin{table}[!htbp]
  \caption{Generative-arm accuracy (\%) by domain, 40 questions per domain.}
  \label{tab:gen_domain_full}
  \centering
  \scriptsize
  \setlength{\tabcolsep}{3.5pt}
  \begin{tabular}{@{}lrrrrr@{}}
    \toprule
    Deployment & \texttt{btm\_ccgt} & \texttt{grid\_queue} & \texttt{retrofit} & \texttt{gulf\_water} & \texttt{permian\_gas} \\
    \midrule
    GPT-OSS-120B (OR) & 77.5 & 80.0 & 70.0 & 65.0 & 65.0 \\
    GPT-OSS-120B (Groq) & 75.0 & 80.0 & 70.0 & 62.5 & 67.5 \\
    GPT-OSS-Safeguard-20B (OR) & 77.5 & 67.5 & 75.0 & 70.0 & 65.0 \\
    GPT-OSS-Safeguard-20B (Groq) & 77.5 & 67.5 & 67.5 & 70.0 & 67.5 \\
    GPT-OSS-20B (OR) & 72.5 & 75.0 & 65.0 & 62.5 & 67.5 \\
    GPT-OSS-20B (Groq) & 72.5 & 62.5 & 70.0 & 55.0 & 67.5 \\
    Qwen3.8-27B (OR) & 75.0 & 62.5 & 70.0 & 45.0 & 62.5 \\
    Qwen3.8-27B (Groq) & 52.5 & 62.5 & 55.0 & 42.5 & 52.5 \\
    Mistral-Small-24B (OR) & 32.5 & 10.0 & 20.0 & 22.5 & 5.0 \\
    Llama-3.1-8B-Instruct (OR) & 10.0 & 7.5 & 10.0 & 17.5 & 25.0 \\
    Mistral-Nemo (OR) & 5.0 & 15.0 & 10.0 & 17.5 & 15.0 \\
    \midrule
    \multicolumn{6}{@{}l}{\emph{Tool-enabled systems (server-side tools on; not open-weight)}}\\
    Groq Compound-mini & 72.5 & 67.5 & 70.0 & 70.0 & 70.0 \\
    Groq Compound & 62.5 & 50.0 & 60.0 & 52.5 & 67.5 \\
    \bottomrule
  \end{tabular}
\end{table}

\begin{table}[!htbp]
  \caption{Multiple-choice accuracy (\%) by skill label, pooled over both banks (questions per skill in the header). Only App, Prog, Reach, and Val share a name with a generative question type, and none is the same task: multiple-choice App asks which stated option complies with a stated limit.}
  \label{tab:mcq_task_full}
  \centering
  \scriptsize
  \setlength{\tabcolsep}{3.5pt}
  \begin{tabular}{@{}lrrrrrrrr@{}}
    \toprule
    Deployment & App (29) & ConsRec (34) & Inv (29) & Prog (38) & Reach (24) & Temp (24) & Trade (41) & Val (21) \\
    \midrule
    GPT-OSS-120B (OR) & 93.1 & 94.1 & 93.1 & 78.9 & 95.8 & 95.8 & 100.0 & 81.0 \\
    GPT-OSS-120B (Groq) & 89.7 & 94.1 & 93.1 & 81.6 & 87.5 & 95.8 & 100.0 & 81.0 \\
    GPT-OSS-Safeguard-20B (OR) & \multicolumn{8}{c}{not run} \\
    GPT-OSS-Safeguard-20B (Groq) & 96.6 & 94.1 & 89.7 & 78.9 & 87.5 & 91.7 & 97.6 & 85.7 \\
    GPT-OSS-20B (OR) & 96.6 & 94.1 & 93.1 & 81.6 & 87.5 & 100.0 & 100.0 & 85.7 \\
    GPT-OSS-20B (Groq) & 93.1 & 94.1 & 93.1 & 84.2 & 83.3 & 95.8 & 100.0 & 85.7 \\
    Qwen3.8-27B (OR) & \multicolumn{8}{c}{not run} \\
    Qwen3.8-27B (Groq) & 93.1 & 100.0 & 55.2 & 63.2 & 95.8 & 66.7 & 80.5 & 81.0 \\
    Mistral-Small-24B (OR) & 86.2 & 94.1 & 48.3 & 76.3 & 95.8 & 75.0 & 82.9 & 81.0 \\
    Llama-3.1-8B-Instruct (OR) & 75.9 & 67.6 & 31.0 & 39.5 & 70.8 & 58.3 & 70.7 & 52.4 \\
    Mistral-Nemo (OR) & 79.3 & 76.5 & 27.6 & 47.4 & 62.5 & 83.3 & 61.0 & 52.4 \\
    \midrule
    \multicolumn{9}{@{}l}{\emph{Tool-enabled systems (server-side tools on; not open-weight)}}\\
    Groq Compound-mini & 89.7 & 97.1 & 93.1 & 92.1 & 91.7 & 95.8 & 97.6 & 85.7 \\
    Groq Compound & 89.7 & 91.2 & 93.1 & 84.2 & 87.5 & 91.7 & 97.6 & 90.5 \\
    \bottomrule
  \end{tabular}
\end{table}

\begin{table}[!htbp]
  \caption{Multiple-choice accuracy (\%) by content bucket, 40 questions per bucket.}
  \label{tab:mcq_domain_full}
  \centering
  \scriptsize
  \setlength{\tabcolsep}{3.5pt}
  \begin{tabular}{@{}lrrrrrr@{}}
    \toprule
    Deployment & Env. & Power & AI wkld. & Fin./siting & Cross & Permit \\
    \midrule
    GPT-OSS-120B (OR) & 92.5 & 97.5 & 85.0 & 97.5 & 85.0 & 92.5 \\
    GPT-OSS-120B (Groq) & 90.0 & 95.0 & 82.5 & 97.5 & 87.5 & 92.5 \\
    GPT-OSS-Safeguard-20B (OR) & \multicolumn{6}{c}{not run} \\
    GPT-OSS-Safeguard-20B (Groq) & 87.5 & 97.5 & 82.5 & 97.5 & 82.5 & 95.0 \\
    GPT-OSS-20B (OR) & 95.0 & 97.5 & 85.0 & 95.0 & 90.0 & 92.5 \\
    GPT-OSS-20B (Groq) & 90.0 & 97.5 & 87.5 & 97.5 & 87.5 & 90.0 \\
    Qwen3.8-27B (OR) & \multicolumn{6}{c}{not run} \\
    Qwen3.8-27B (Groq) & 75.0 & 80.0 & 75.0 & 75.0 & 72.5 & 97.5 \\
    Mistral-Small-24B (OR) & 72.5 & 85.0 & 62.5 & 85.0 & 75.0 & 100.0 \\
    Llama-3.1-8B-Instruct (OR) & 45.0 & 57.5 & 47.5 & 57.5 & 57.5 & 85.0 \\
    Mistral-Nemo (OR) & 65.0 & 60.0 & 40.0 & 60.0 & 52.5 & 87.5 \\
    \midrule
    \multicolumn{7}{@{}l}{\emph{Tool-enabled systems (server-side tools on; not open-weight)}}\\
    Groq Compound-mini & 95.0 & 97.5 & 87.5 & 100.0 & 87.5 & 92.5 \\
    Groq Compound & 92.5 & 95.0 & 80.0 & 97.5 & 85.0 & 95.0 \\
    \bottomrule
  \end{tabular}
\end{table}

\begin{table}[!htbp]
  \caption{Listing every applicable action (generative \textsc{App}, 25 questions). Strict: exact set match, as the prompt asks (an action whose preconditions hold counts even if it changes nothing). Excl.\ no-op: exact match after removing such no-op actions from both the gold set and the answer. P, R, F1, J: mean precision, recall, F1, and Jaccard overlap. J$\geq$.8: share of questions with Jaccard at least 0.8. R chg./R no-op: recall on gold actions that change the state, and on no-op ones. Omit.\ no-op: share of all omitted gold actions that are no-ops. 66\% of gold actions are no-ops. The last row lists every grounded action without reading the state: overlap metrics reward it, exact match does not.}
  \label{tab:app_full}
  \centering
  \scriptsize
  \setlength{\tabcolsep}{3.5pt}
  \begin{tabular}{@{}lrrrrrrrrrr@{}}
    \toprule
    Deployment & Strict & Excl. & P & R & F1 & J & J$\geq$.8 & R & R & Omit. \\
    & & no-op & & & & & & chg. & no-op & no-op (\%) \\
    \midrule
    GPT-OSS-120B (OR) & 0.0 & 0.0 & 0.94 & 0.37 & 0.50 & 0.35 & 4.0 & 0.78 & 0.16 & 88 \\
    GPT-OSS-120B (Groq) & 0.0 & 4.0 & 0.89 & 0.36 & 0.48 & 0.35 & 12.0 & 0.72 & 0.17 & 85 \\
    GPT-OSS-Safeguard-20B (OR) & 0.0 & 24.0 & 0.91 & 0.44 & 0.57 & 0.42 & 4.0 & 0.85 & 0.21 & 91 \\
    GPT-OSS-Safeguard-20B (Groq) & 0.0 & 20.0 & 0.89 & 0.39 & 0.51 & 0.36 & 0.0 & 0.82 & 0.15 & 90 \\
    GPT-OSS-20B (OR) & 0.0 & 16.0 & 0.84 & 0.45 & 0.55 & 0.41 & 0.0 & 0.82 & 0.24 & 89 \\
    GPT-OSS-20B (Groq) & 0.0 & 24.0 & 0.89 & 0.50 & 0.61 & 0.47 & 12.0 & 0.85 & 0.31 & 89 \\
    Qwen3.8-27B (OR) & 4.0 & 8.0 & 0.67 & 0.30 & 0.37 & 0.25 & 4.0 & 0.57 & 0.16 & 79 \\
    Qwen3.8-27B (Groq) & 0.0 & 0.0 & 0.69 & 0.30 & 0.37 & 0.24 & 0.0 & 0.54 & 0.17 & 78 \\
    Mistral-Small-24B (OR) & 0.0 & 0.0 & 0.73 & 0.15 & 0.23 & 0.14 & 0.0 & 0.39 & 0.01 & 76 \\
    Llama-3.1-8B-Instruct (OR) & 0.0 & 4.0 & 0.65 & 0.56 & 0.52 & 0.38 & 0.0 & 0.65 & 0.52 & 70 \\
    Mistral-Nemo (OR) & 0.0 & 0.0 & 0.73 & 0.11 & 0.18 & 0.10 & 0.0 & 0.23 & 0.04 & 71 \\
    \midrule
    \multicolumn{11}{@{}l}{\emph{Tool-enabled systems (server-side tools on; not open-weight)}}\\
    Groq Compound-mini & 0.0 & 8.0 & 0.88 & 0.33 & 0.46 & 0.32 & 4.0 & 0.76 & 0.09 & 88 \\
    Groq Compound & 4.0 & 8.0 & 0.85 & 0.44 & 0.53 & 0.40 & 8.0 & 0.80 & 0.24 & 89 \\
    \midrule
    List every grounded action & 0.0 & 0.0 & 0.56 & 1.00 & 0.71 & 0.56 & 0.0 & 1.00 & 1.00 & -- \\
    \bottomrule
  \end{tabular}
\end{table}

\begin{table}[!htbp]
  \caption{What happened to each of the 200 generative answers. Wrong: a parsed answer that names only objects in the world but fails the validator. Ungrounded: names a fact or action that does not exist in the world (or an index past the plan). Unparseable: no answer of the requested shape. Out of tokens: the response hit its token limit. Last column, for the 25 \textsc{App} answers that missed: omissions only / omissions and extra actions / extra actions only.}
  \label{tab:outcomes}
  \centering
  \scriptsize
  \setlength{\tabcolsep}{3.5pt}
  \begin{tabular}{@{}lrrrrrrc@{}}
    \toprule
    Deployment & Correct & Wrong & Ungrounded & Unparseable & Out of tokens & Empty & \textsc{App} misses \\
    \midrule
    GPT-OSS-120B (OR) & 143 & 56 & 0 & 1 & 0 & 0 & 15/9/1 \\
    GPT-OSS-120B (Groq) & 142 & 58 & 0 & 0 & 0 & 0 & 12/12/1 \\
    GPT-OSS-Safeguard-20B (OR) & 142 & 57 & 1 & 0 & 0 & 0 & 12/13/0 \\
    GPT-OSS-Safeguard-20B (Groq) & 140 & 58 & 2 & 0 & 0 & 0 & 12/13/0 \\
    GPT-OSS-20B (OR) & 137 & 59 & 1 & 0 & 3 & 0 & 13/12/0 \\
    GPT-OSS-20B (Groq) & 131 & 69 & 0 & 0 & 0 & 0 & 12/12/1 \\
    Qwen3.8-27B (OR) & 126 & 69 & 3 & 1 & 1 & 0 & 6/18/0 \\
    Qwen3.8-27B (Groq) & 106 & 90 & 3 & 1 & 0 & 0 & 6/19/0 \\
    Mistral-Small-24B (OR) & 36 & 144 & 20 & 0 & 0 & 0 & 11/14/0 \\
    Llama-3.1-8B-Instruct (OR) & 28 & 156 & 13 & 3 & 0 & 0 & 5/19/1 \\
    Mistral-Nemo (OR) & 25 & 127 & 48 & 0 & 0 & 0 & 12/13/0 \\
    \midrule
    \multicolumn{8}{@{}l}{\emph{Tool-enabled systems (server-side tools on; not open-weight)}}\\
    Groq Compound-mini & 140 & 59 & 1 & 0 & 0 & 0 & 12/13/0 \\
    Groq Compound & 117 & 72 & 11 & 0 & 0 & 0 & 10/13/1 \\
    \bottomrule
  \end{tabular}
\end{table}

\begin{table}[!htbp]
  \caption{Inference settings and cost. Every call used temperature 0 and one attempt per question, re-asked only after a provider error; the generative arm prompts with two worked examples from a held-out tutorial domain, the multiple-choice arm with none. Temperature 0 was not deterministic on these APIs: answers that exhausted their budget and were re-asked came back answered. Tokens in millions (prompt / completion). Cost in US\$, both arms, as reported by OpenRouter (total \$2.00). Tool rows: answers whose trace shows a server-side tool call (generative / multiple choice). $^{a}$Or the model's provider cap if lower. $^{b}$Groq reports tokens but not cost. $^{c}$The harness sets \code{reasoning\_effort=none} for Qwen on Groq and caps its output at 2,048 tokens; OpenRouter leaves reasoning on.}
  \label{tab:settings}
  \centering
  \scriptsize
  \setlength{\tabcolsep}{3.5pt}
  \begin{tabular}{@{}l l r r r r r@{}}
    \toprule
    Deployment & Reasoning & Max out & Gen tokens & MCQ tokens & Cost & Tool rows \\
    \midrule
    GPT-OSS-120B (OR) & default & 16,384$^{a}$ & 0.40 / 0.24 & 0.07 / 0.07 & 0.14 & 0 / 0 \\
    GPT-OSS-120B (Groq) & default & 16,384$^{a}$ & 0.40 / 0.22 & 0.07 / 0.06 & --$^{b}$ & 0 / 0 \\
    GPT-OSS-Safeguard-20B (OR) & default & 16,384$^{a}$ & 0.40 / 0.42 & -- & 0.15 & 0 / -- \\
    GPT-OSS-Safeguard-20B (Groq) & default & 16,384$^{a}$ & 0.40 / 0.39 & 0.07 / 0.11 & --$^{b}$ & 0 / 0 \\
    GPT-OSS-20B (OR) & default & 16,384$^{a}$ & 0.40 / 0.41 & 0.07 / 0.11 & 0.09 & 0 / 0 \\
    GPT-OSS-20B (Groq) & default & 16,384$^{a}$ & 0.40 / 0.39 & 0.07 / 0.14 & --$^{b}$ & 0 / 0 \\
    Qwen3.8-27B (OR) & on & 16,384$^{a}$ & 0.42 / 0.49 & -- & 1.57 & 0 / -- \\
    Qwen3.8-27B (Groq) & off$^{c}$ & 2,048 & 0.40 / 0.01 & 0.06 / 0.00 & --$^{b}$ & 0 / 0 \\
    Mistral-Small-24B (OR) & n/a & 16,384$^{a}$ & 0.44 / 0.01 & 0.10 / 0.00 & 0.03 & 0 / 0 \\
    Llama-3.1-8B-Instruct (OR) & n/a & 16,384$^{a}$ & 0.40 / 0.01 & 0.06 / 0.00 & 0.01 & 0 / 0 \\
    Mistral-Nemo (OR) & n/a & 16,384$^{a}$ & 0.41 / 0.00 & 0.06 / 0.00 & 0.02 & 0 / 0 \\
    \midrule
    \multicolumn{7}{@{}l}{\emph{Tool-enabled systems (server-side tools on; not open-weight)}}\\
    Groq Compound-mini & default & 16,384$^{a}$ & 0.87 / 0.25 & 0.22 / 0.07 & --$^{b}$ & 1 / 3 \\
    Groq Compound & default & 16,384$^{a}$ & 2.09 / 0.29 & 0.67 / 0.13 & --$^{b}$ & 128 / 82 \\
    \bottomrule
  \end{tabular}
\end{table}

\begin{table}[!htbp]
  \caption{The five generative-arm domains. Means over generated instances; each domain contributes 40 questions, five per type. Applicable: mean actions applicable in a sampled state, counting no-ops.}
  \label{tab:domains}
  \centering
  \scriptsize
  \setlength{\tabcolsep}{3.5pt}
  \begin{tabular}{@{}l p{2.4cm} p{3.6cm} rrrrr@{}}
    \toprule
    Domain & Models & What binds & Facts & Actions & Applic. & Plan & States \\
    \midrule
    \texttt{btm\_ccgt} & Campus on its own gas turbines & Air permit, gas lateral, and fuel contract before any unit is energized & 37.0 & 32.5 & 17.2 & 17.6 & 65,911 \\
    \texttt{grid\_queue} & Utility interconnection & Screening, queue, phase-1 study, agreement, and security before any build; the three-to-seven-year path & 35.5 & 35.0 & 14.0 & 15.8 & 7,362 \\
    \texttt{retrofit} & Densifying a live site & Feeder budget: a dense rack costs two feeders air-cooled, one liquid-cooled & 34.2 & 31.2 & 12.4 & 9.9 & 6,946 \\
    \texttt{gulf\_water} & Coastal site, scarce water & One-shot cooling fork, evaporative or dry; sets the site's water draw & 21.6 & 21.5 & 8.9 & 10.4 & 241 \\
    \texttt{permian\_gas} & Gensets at well pads & Lease and gas-quality statics decide which pads can host a unit & 36.9 & 50.6 & 11.8 & 11.9 & 45,433 \\
    \bottomrule
  \end{tabular}
\end{table}

\end{document}